# The Online Discourse of Virtual Reality and Anxiety

Kwabena Yamoah[1] and Cass Dykeman[1]
[1]Oregon State University

Kwabena Yamoah  https://orcid.org/0009-0004-7784-1593
Cass Dykeman  https://orcid.org/0000-0001-7708-1409

## Abstract

One cannot ignore the impact of technological advancement in virtual reality (VR) in the treatment of clinical concerns such as generalized anxiety disorder or social anxiety. VR has created additional pathways to support patient well-being and care. Understanding online discussion of what users think about this technology may further support its efficacy. The purpose of this study was to employ a corpus linguistic methodology to identify the words and word networks that shed light on the online discussion of virtual reality and anxiety. Using corpus linguistics, frequently used words in discussion along with collocation were identified by utilizing Sketch Engine software. The results of the study, based upon the English Trends corpus, identified "VR," "Oculus," and "headset" as the most frequently discussed within the VR and anxiety subcorpus. These results point to the development of the virtual system, along with the physical apparatus that makes viewing and engaging with the virtual environment possible. Additional results point to collocation of prepositional phrases such as "of virtual reality," "in virtual reality," and "for virtual reality" relating to the design, experience, and development, respectively. These findings offer new perspective on how VR and anxiety together are discussed in general discourse and offer pathways for future opportunities to support counseling needs through development and accessibility.

## Introduction

Anxiety disorders are important to study for the development of potential treatments. Due to unique features and common co-occurring disorders, such as panic, obsessive compulsive, and posttraumatic stress disorder, Alomari et al. (2022) researched anxiety disorders and determined better multimodal support for the millions of individuals who experience anxiety to increase overall quality of life. Disorders should be studied for other reasons such as identifying maladaptive versus appropriate coping skills for disorders. Hofmann (2007) highlighted that individuals with maladaptive coping skills related to perceived social standards lacked defined social goals, critical focus on attention, negative self-perception, rumination, and poor social skills that maintain the cycle of a negative view of self in context with others. These issues could possibly be dealt with by employing the use of tools within cognitive behavior therapy (CBT).

**Rationale**

Anxiety disorders should be studied to assess the impact of therapeutic models such CBT on anxiety symptoms and diagnosis. For instance, in a meta study on anxiety, Stewart and Chambless (2009) argued for further exploration of CBT tools on anxiety symptoms over time to offer greater insight into the durability of treatment in various settings.

The utility of VR in counseling should be studied to identify advancements in methodology and utility. Anderson et al. (2013) highlighted positive effects of VR on social fears, while Maples-Keller et al. (2017) identified appropriate candidates for VR clinical care. Several rationales also exist for studying the use of VR in treating anxiety disorders. First, McMahon (2025) shared that VR can create immersive environments that not only curate emotional experiences but can enhance therapeutic outcomes. VR is adaptive, can meet the needs of diverse clients, and is effective in treating various anxiety disorders. Another rationale for studying the use of VR in treating anxiety disorders is the assessment and identification of anxiety disorders. In a review of games and VR for assessing depression and anxiety, Chitale et al. (2022) found utility in VR environments built to simulate anxiety-producing scenarios within a controlled setting, thus identifying anxiety levels and triggers for individuals. These authors also noted that studying the use of VR in treating anxiety disorders helps to identify treatment efficacy.

Baghaei et al. (2021) found that VR exposure therapy is effective in treating anxiety disorders by offering safe and controlled spaces. Individuals can practice and navigate through scenarios within the safety of clinical environments. VR can also serve as adjunctive therapy within medical settings, further solidifying its rationale. In a study focusing on breast cancer patients, Heyll (2022) found positive benefits of VR for anxiety and depression. Heyll (2022) found statistical significance emanating from a pre and posttest on chemotherapy anxiety and depression, with VR improving quality of life and treatment outcome for participants. Finally, the use of VR in treating anxiety disorders can be advantageous over traditional counseling methods and theory. Gaggioli et al. (2014), in research on experiential scenarios and psychological distress, identified reduction in anxiety symptoms for participants who utilized CBT in conjunction with VR exposure. These findings demonstrate the wide and substantive impact of VR on reduction of anxiety disorders over traditional options.

**Relevant Background**

In preparation for the research question development, four areas were explored. These were (a) key definitions, (b) what is known about anxiety, (c) what is known about VR as a treatment modality for anxiety, and (d) what is known about the word usage patterns of anxiety web-based discussions within a community using keyness and collocation.

A couple key terms are used in the study. Maulana and Purnomo (2021) defined *virtual reality* as a computer-simulated environment that allows users to digitally interact with a constructed universe that mimics real-world settings. This immersive interface is designed to replicate actual surroundings. In this type of setting, individuals can engage in activities and environments and participate in learning outcomes that allow them to gain a greater sense of themselves and a truly unique experience of reality. Donnelly et al. (2021) indicated the utilization of VR is a valuable tool to create environments in which individuals can navigate and participate in simulated settings as treatment options for their mental health.

The second key term is *anxiety disorders*, defined in the *Diagnostic and Statistical Manual of Mental Disorders* as characterized with commonalities inclusive of excessive fear and related behavioral disturbances that has significant impact on the well-being of individuals (American Psychiatric Association, 2022). Anxiety disorders include separation anxiety disorder, panic disorder, agoraphobia, and generalized anxiety disorder, among others. Szuhany and Simon (2022) shared that anxiety is the second most common mental health disorder after depression. Additionally, anxiety disorders cost the U.S. economy over 46 billion dollars in the form of workplace costs (Harder et al., 2016). This indicates the widespread financial impact of these disorders, as well as the challenges faced by work environments in managing the effects of anxiety disorders among employees.

Within the review of the literature, the hallmark features of this disorder are excessive fear and related behavioral disturbances. For many, anxiety manifests as self-consciousness or even perceived severe judgment from self and others. Such feelings lead people with anxiety disorders to feel stressed. Weeks (2014) shared that those with anxiety disorders overestimate their negative evaluations on their performance, leading to faulty appraisals, distortions, and predications of consequences of failure in social settings. Thus, they engage in hypervigilance, characterized by a heightened state of awareness and scanning of the environment for potential threats. This can lead to excessive monitoring of one's own behavior and circumstances to avoid being perceived as lacking in something.

Research on anxiety disorders reveals that there are multifaceted influences with regards to origins, development, presentation, intensity, and duration. Penninx et al. (2021) highlighted several influential pieces related to the dynamics of the disorder including (a) core emotions and thoughts such as excessive and concerning fear; (b) physical symptoms inclusive of restlessness, fatigue, irritability, difficulty concentrating; (c) avoidance of fear-inducing situations; (d) presentation and duration from 1–6 months for diagnosis; and (e) age of symptom onset for an anxiety disorder, such as selective mutism beginning in childhood, to generalized anxiety disorder in adulthood. These attributes extend the life cycle of anxiety disorders, impacting the daily functioning of those affected.

Barkowski et al. (2020) reviewed the effectiveness of group psychotherapy in the treatment of anxiety disorders. The results of the meta-analysis revealed that group psychotherapy had a significant impact on the treatment of anxiety disorders and reduced symptoms more effectively than control groups without treatment. However, there were no significant differences when comparing the group results to those engaged in psychotherapy or pharmacotherapy (Barkowski et al., 2020).

In research highlighting social networks and individuals with anxiety disorders, Nguyen et al. (2024) identified that lifestyle impacts those diagnosed with anxiety disorders, such as subjective, interpersonal, and structural isolation. Structural isolation, especially being childless, was negatively associated as a contributing factor to agoraphobia. To assist those experiencing social isolation, the researchers argued for greater public health support and program implementation to alleviate the burdens of anxiety disorders. This is significant as one considers supportive mechanisms to enhance functional living for individuals diagnosed with anxiety disorders who have the need for social connection. This could have significant implications on health and behavioral systems.

Research also shows significant differences between genders when coping with the effects of anxiety disorders. Vanderminden and Esala (2019) found that even after controlling for anxiety symptoms, women were more prone to being diagnosed with an anxiety disorder compared to men. This finding offers insights into potential biases in the diagnosis of anxiety disorders based on gender. As such, there may be differences in treatment for men and women.

Despite more recent knowledge of VR, its implications in counseling, and its effectiveness with anxiety disorders, little is known about perspectives on the utilization and adoption by clinicians. Rimer et al. (2021), in their study of VR exposure therapy, highlighted the concern of some clinicians regarding wide-scale implementation and the difficulties of hardware and software adoption. Future research is needed to ensure the viability of methods to treat various anxiety disorders and also implementations costs, especially with utility in clinical settings.

Donker et al. (2019), researching acrophobia and the use of a self-guided VR app based on CBT principles, advocated for more research in VR. They drew conclusions concerning the effectiveness of the modality, finding that the VR app led to significant decreases in acrophobia symptoms. By performing additional studies, more can be known about how to apply VR to anxiety disorders and help identify methods to support individuals challenged with anxiety.

When attempting to piece together a narrative on VR and anxiety disorders concerning keywords and collocation, current research is limited on this topic. However, in considering similar research methods, several studies identified the importance of looking at keyness and collocation in research. Hamed (2020) described utilizing keyness to identify thematic expressions within content covering U.S. presidents. Hamed found that keywords identified showed differences in administrations and political parties by either an internal or external political issues. Collocation, according to Hamad, interpret how particular words are regularly aligned together within the span of a central word. These two parts working independently and collectively at the same time help interpret the relationship amongst words used and messaging.

Similarly, Chua et al. (2023) applied keyness and collocation in their study on domestic violence. They found that words on Twitter underscored the context, count, and association of words used together around significant terminology for domestic violence. These two depictions offered by Hamed (2023) and Chua et al. illustrate the possibilities of depicting not only significant words concerning VR and social anxiety but also the frequency of utilization.

**Statement of Research Questions**

Given the aforementioned, the following research questions were developed to guide the present study:

RQ1: In reference to the general online discourse, what words distinguish the web-based discourse on VR and anxiety?

RQ2: Within the online discourse on VR and anxiety, what is the word network surrounding the node bigram "virtual reality?"

RQ3: Within the online discourse on VR and anxiety, what is the word network surrounding the first-letter initialism "VR?"

**Method**

**Design**

With the study focusing on the discourse of VR with persons with anxiety, the following method was employed to address the research questions. Corpus linguistic design methodology (Weisser, 2016) was used to identify and incorporate four variables: corpus, keyness, node, and colocation. Corpus linguistics makes use of considerable amounts of both spoken and written natural texts, to identify insights with language (Reppen & Simpson-Vlach, 2019). The corpus is comprised of archived discussions in Sketch Engine (2024a), which allows for natural language processing, computational linguistics, and electronic lexicography. English Trends corpus, identified as a Sketch Engine monitor corpus, consists of news, Wikipedia articles, and texts from sources that publish daily updates through web feeds that can capture discussion and content on a specific topic, such as VR and anxiety. Sketch Engine encompasses conversations of topics. The level of measure for keyness and collocation was continuous and for corpus and node was nominal. The unit of analysis was individual words.

**Study Corpus**

*Register, Scope, and Source*

The register for this corpus was a subcorpus of online news articles and media discourse titled "Virtual Reality, VR, and Anxiety" on the https://www.sketchengine.eu website. The scope included discussions from 2014 to present. The source was the English Trends corpus (Sketch Engine, 2024a) available within the online corpus linguistics software Sketch Engine (Sketch Engine, 2024c). The corpus shared a broad array of information and discourse from internet sources, as the subcorpus presents a more focused discussion concerning VR and anxiety.

*Preprocessing*

The corpus is readily accessible through SketchEngine (2024a; 2024b; 2024c) and requires no additional preprocessing.

**Reference Corpus**

*Register, Scope, and Sources*

The register was online news articles and media discourse. The source was Sketch Engine. The English Trends corpus comprised the text utilized for the study (Sketch Engine, 2024a). English trends contains 86 billion words, nearly 100 billion tokens taken from newsfeeds. The scope was the same as for study corpus. The source was the whole English Trends corpus except for the documents that were part of the study corpus.

*Preprocessing*

Given the preprocessing done during the creation of the English trends (Sketch Engine, 2024a), no additional processing was needed.

**Measures**

*Keyness*

Keyness denotes words that frequently appear in a text corpus (Gabrielatos, 2018). Analyzing these keywords can uncover the themes within the discourse of a group. This analysis encompasses not only the frequency of word occurrences but also their dispersion, or the distribution of word usage across the corpus (Egbert & Biber, 2019).

*Word Node*

The node is the central word or phrase connected to the collocates.

*Bigram*

A bigram is the two-word form of an n-gram. N-grams are also referred to as multiword units. For the present study, the bigram "virtual reality" served as the node for RQ2.

*Initialism*

An initialism is an abbreviation created by substituting initial letters of a group of words for the individual words (Bloom, 2000). In terms of the present study, VR is a first-letter initialism for the bigram "virtual reality" and serves as the node for RQ3.

*Collocation*

This term refers to the process of categorizing words that are frequently used near a particular word or phrase (Baker et al., 2006; Sketch Engine, 2024b). This process helps researchers understand the role of specific words in the discourse of a group through dimensions such as proximity, frequency, and context.

**Apparatus**

SketchEngine (2024c) was the software utilized for analyzing content. It is an online analysis platform for text and is designed to work with wide-ranging text corpora. It detects and categorizes distinct language and patterns of language frequently, as well as infrequent or obsolete words and patterns of one's grammatical constructs. By analyzing these trends, Sketch Engine can offer insights into how language progresses and develops through time.

**Data Analysis**

For RQ1 (keyness), the following descriptive statistics are provided: (a) raw count, (b) normalized frequency (count per million), and (c) relative document frequency (percentage). In addition, an effect size of the differences between two corpuses are reported using Simple Maths (Kilgarriff, 2009). Regarding RQs 2–3 (collocation), the following are conveyed: collocate, collocate frequency, and grammatical relationship. The strength of the association between the node of collocated was assessed via logDice (Rychlý, 2009).

**Results**

With RQ1 (words that are used frequently in the discourse of VR and anxiety), the three strongest words were "VR," "Oculus," and "headset." A listing of the top 15 keywords displayed in the Virtual Reality, VR, and Anxiety subcorpus can be found in Table 3.1.

RQ2 looked at the collocation and word association with the bigram "virtual reality." The top three words aligned with modifiers of VR were "location-based," "immersive," and "Rift." The top three words aligned with nouns modified by VR were "VR," "Bavor" and "pov." Finally the top three collocated of VR with prepositional phrases included "of virtual reality," "in virtual reality," and for "virtual reality." The top 15 collocation review aligned with the bigram "virtual reality" can be found in Table 3.2. Finally, a list of the top 15 collocates associated with VR can be viewed as part of a visual with the bigram "virtual reality" in the center of Figure 3.1.

RQ3 looked at the collocation and word association with the initialism VR. The top three words aligned with modifiers of VR were "Gear," "PlayStation," and "Oculus." The top three words aligned with nouns modified by VR were "headset," "goggle," and "AR." The top three collocates of the initialism VR with an and/or are "AR," "Rift," and "Vive." Finally, the top three collocated of VR with prepositional phrases included "in VR," "of VR," and "for VR." The top 15 collocations review aligned with the initialism VR can be found in Table 3.3. Finally, a list of the top 15 collocates associated with VR can be viewed as part of a visual with the initialism VR in the center of Figure 3.2.

**Discussion**

The study examined the online discourse and word network involving virtual reality, VR, and anxiety.

**RQ1 Keyness**

The underpinning of RQ1 was to understand the words that distinguished the language on VR and anxiety to gain greater perspective and frequency of use within discussions. The results revealed that the most used keywords surrounding the web-based discourse on VR and anxiety were "VR," "Oculus," and "headset." The identification of these terms indicates that types of apparatuses used to deploy virtual immersive environments can be central in healthcare, education, and research. Moro et al. (2017) found that the use of cost effective mobile based virtual reality headsets in medical and health care education was as effective as desktop based VR systems. Craig et al. (2022) examined the Oculus Quest headset for VR, finding that the apparatus was effective at detecting changes in postural control as a result of different types of visual field manipulations and thus was an alternative to lab based balance assessments within healthcare. When considering anxiety, Bennett and Christensen (2024) studied the use of the Oculus headset and found a positive impact for reducing anxiety in college students. It is plausible these results signal the accessibility of VR technology for different populations and the need to support the clinical needs of users who face anxiety disorders (Oing & Prescott, 2018). Graham et al. (2025) asserted that there is the need for increased technology for self-guided VR experiences as it is an engaging approach for effectively and safely treating anxiety disorders. Singh (2025) argued that there is a need for advancement in technology due to device weights leading to user physical and mental fatigue. Based on advances in technology, improvements to VR technology such as headsets would add value to user experiences.

**RQ2 Collocates of Virtual Reality**

RQ2 explored the online conversations and collocations of words associated with the bigram "virtual reality." Analysis of the results revealed the top three commonly associated modifiers with VR, top three noun modifiers with VR, and top three prepositional phrases with VR.

*Words Modified by Virtual Reality*

The top three collocates commonly associated with words modified by VR were "location-based," "immersive," and "Rift." These identify the structure of the virtual environment with respect to being in the physical environment, along with the device use—in the Oculus Rift, a headset device developed by Meta for immersive viewing (Iriye & Jacques, 2021). "Location-based" could also appear due to the ability of an engaged user—from their current position, viewing content that is enhanced by the VR environment. "Immersive" and "Rift" may also be used as one considers the adoption of a curated virtual environment that is made possible by a headset, such as Rift, that supports the immersive software. Another perspective on these results could be the ability of the virtual environment to provide realism to the user experience as the user navigates through an anxiety reduction perspective. This is plausible as research from Malbos et al. (2025) highlighted the efficacy of virtual reality by users practicing within a technological environment to treat generalized anxiety disorder.

*Collocated With Virtual Reality as a Noun*

The three top words collocated with VR as a noun were "VR," "Bavor," and "pov." This could illuminate the discussion around VR encompassing a known leader in the field of VR, Clay Bavor, who at one point served as Google's executive and technical lead in augmented reality (AR) and VR. His work in the field included immersive technologies such as VR headsets as well as experiences.

"Pov" stands for the user's "point of view," referring to the ability of a VR device to capture the journey of the user. These three words can also point to Mr. Bavor as a pioneer in the space of VR and individual user experiences. This perspective is the most plausible as Mr. Bover spent time at Google shaping the landscape for VR experiences.

*Collocated With Virtual Reality as Prepositional Phrases*

The top three prepositional phrases collocated with VR included "of virtual reality," "in virtual reality," and "for virtual reality." When considering "of virtual reality," users may be referencing the design and ability related to the virtual environment. The structure of "in virtual reality" may reference a user's direct experience of becoming immersed within the environment. This could be inclusive of learning about the particular VR setting as well as the expected experience of VR. Finally, the structure of "for virtual reality" may relate to both the hardware capacity, such as the headset, and the software development of the immersive environment. Combined, these perspectives may lend themselves to the acceptability and unacceptable aspects of VR. As noted by Oing and Prescott (2018), there is a general agreement regarding the viability of VR exposure therapy to support the needs of users. The latter explanation offers greater credence as research and feedback from users shows that VR care is generally accepted in clinical practice to treat anxiety (Oing and Prescott, 2018).

*RQ3 Collocates of VR*

RQ3 explored the online conversations and collocations of words associated with the initialism "VR." Analysis revealed the top three commonly associated modifiers with VR, the top three noun modifiers with VR, the top three collocates of the initialism VR with an and/or are, and the top three prepositional phrases with VR.

*Words Modified by VR*

The top three collocates commonly associated with words modified by VR were "Gear," "PlayStation," and "Oculus." These words could identify the different devices that operate as unique platforms for VR experiences. Gear VR is constructed by Samsung, Oculus Rift by Meta, and PlayStation VR by Sony (Egan et al., 2016; Moro et al., 2017). They also could indicate the first foray into each company's technological advancement with producing VR environments for the user base. Finally, these words may identify the approach taken by each device developer to apply technology to health-related VR and to education experiences. The second explanation provides a compelling reason for these results. These VR devices have proprietary technology that presents a unique platform for each company as they cater to individual users.

*Collocated With VR as a Noun*

The three top words collocated with VR as a noun were "headset," "goggle," and "AR." This could illuminate the discussion around the VR device experience a user has with the structure of the headset or head mounted display, the view through the goggle, and the AR environment. These could also reflect discussion surrounding the key differences among manufacturers of VR devices including their development of the immersive environment (Innocente et al., 2023). The latter argument provides a more plausible interpretation, especially considering the way a manufacturer chooses to develop their VR environment to support anxiety-reduction simulations.

*Collocated With VR and/or*

The top three collocates of the initialism VR with an and/or were "AR," "Rift," and "Vive," which reveals discussions on the devices and their application to a real-world environment such as AR or a digital environment such as VR (Huang et al., 2019). When referencing "and/or" with VR, the discussion is typically categorizing it with manufacturers such as the Oculus Rift and HTC Vive, both of which can overlay within the digital environment of VR. There could be discourse about the benefits and drawbacks of each. Also, VR is separate from how AR is used, and the possibility of the mixed environments of VR and AR that these devices employ within their technology (Oyelere et al., 2020). Finally, there could be discussion on which device and environment best suits the needs of healthcare settings to support the user/patient experience. The latter describes an appropriate explanation of the discussion as selecting the right device can be impactful for a variety of treatments (Kanschik et al., 2023; Tan et al., 2022).

*Collocated With VR with Prepositional Phrases*

The top three words collocated with VR with prepositional phrases included "in VR," "of VR," and "for VR." The structure of "in VR" may reference a user's direct experience in the VR activity and the environment. Users "in VR" can explore a reconstructed digital event and engage with simulations for health-related practices inclusive of patient education (Kanschik et al., 2023). When considering "of VR," there could be discussion on the impact of the VR ecosystem supporting treatment outcomes for anxiety and other phobias (Freeman et al., 2017). Finally, the structure of "for VR" may relate to the unique focused applications of VR in healthcare, education, and general user experiences (Kanschik et al., 2023; Schiza et al., 2019). The argument "for VR" is most compelling because it explains how VR can be used for health care settings and educational facilities. This could lead to future expansion within VR and a targeted approach for developers.

In summary, the outcomes of this study illuminate how keywords identified such as "VR," "Bavor," and "pov" are part of the natural discourse surrounding VR. Also, the outcomes of this study show the ability to identify the word network surrounding "virtual reality" and "VR," and the interpretations of the words that present logical arguments.

**Limitations**

There are several limitations to the study. First, the study data were aggregated from Sketch Engine's English Trends corpus, which included data extracted from January 2014 to September 2025, and was comprised of news, Wikipedia articles, and texts from web feeds. A differing corpus, such as entenTen21, which is a smaller corpus consisting of 52 billion words, could have been used and could have produced different keyness and word networks. Additional filtering could be conducted on the entenTen21 corpus, eliminating poor quality text and spam; thus, the outcome may have differed. Second, though Sketch Engine offers readily available corpora to study, other corpora exist that would potentially provide a different data set and outcomes.

A final limitation is the broad focus of the English Trends corpus. If presented with a corpus focused specifically on anxiety, healthcare, technology, education, neuroscience, or even social sciences, the results may have been different from what was presented, as each topic has a unique focus and engaged population of data to be evaluated.

**Implications Practice and Research**

When considering the results and discussion from this study, there are two implications for future research. First, there is an opportunity to identify standardization of VR technology with consideration of anxiety, along with the appropriate requirements for a headset and goggle. As the frequents of keywords "Oculus" and "headset" were indicated, there is the recommendation to research the specifications of what users deem appropriate for VR devices. For example, clinical supervisors may need specific protocols on how to support their clinicians to help users with technology. If there are standard processes to consider, the onboarding and training for the devices may be streamlined. Currently, there is research on the efficacy of VR for treatment of anxiety (Graham et al., 2025), but little discussion exists on the standardization for VR protocols that would be applicable to anxiety settings and supervisor, clinician, and user needs. The standardization may present opportunities for greater commercialization of VR technology for anxiety.

An additional implication reflects the appropriate experience and safety of the device for utilization. Future research should review the side effects of engaging within a VR environment for anxiety purposes. One must view the potential benefits of treatment versus possible psychological drawback (Jingili et al., 2023). This can be influential in the types of devices selected for VR and the amount of time using them. A final implication is the acceptance of VR as a readily accessible instrument for assessing and treating patients in clinical practices.

**Table 3.1**

*Top 15 Words With Strong Keyness (RQ1)*

| | Lemma | Frequency Focus | Frequency Reference | Frequency per million Focus | Frequency per million Reference | Document frequency Focus | Document frequency Reference | Relative DOCF Focus | Relative DOCF Reference | ARF Focus | ARF Reference | ALDF Focus | ALDF Reference | Score |
|---|---|---|---|---|---|---|---|---|---|---|---|---|---|---|
| 1 | vr | 1,249,296 | 1,489,295 | 36,026.90 | 14.76 | 15,613 | 427,969 | < 0.01% | 0.16% | 931,354.00 | 346,311.75 | 987,293.31 | 240,546.45 | 2,286.3 |
| 2 | oculus | 102,838 | 332,776 | 2,965.62 | 3.30 | 11,707 | 121,637 | < 0.01% | 0.05% | 47,098.71 | 89,635.89 | 41,896.32 | 62,339.22 | 690.3 |
| 3 | headset | 198,515 | 965,740 | 5,724.73 | 9.57 | 13,605 | 357,118 | < 0.01% | 0.14% | 105,293.95 | 278,995.41 | 100,941.84 | 210,165.84 | 541.7 |
| 4 | vive | 38,766 | 129,655 | 1,117.92 | 1.28 | 8,175 | 54,426 | < 0.01% | 0.02% | 17,396.65 | 38,223.38 | 13,861.48 | 25,972.70 | 489.7 |
| 5 | ar | 120,931 | 819,715 | 3,487.38 | 8.12 | 12,209 | 363,772 | < 0.01% | 0.14% | 51,593.23 | 278,957.88 | 41,080.10 | 216,472.33 | 382.4 |
| 6 | htc | 38,382 | 405,414 | 1,106.85 | 4.02 | 8,506 | 126,740 | < 0.01% | 0.05% | 17,828.42 | 89,843.38 | 14,638.62 | 49,153.12 | 220.8 |
| 7 | daydream | 16,642 | 139,180 | 479.92 | 1.38 | 3,850 | 100,131 | < 0.01% | 0.04% | 5,054.91 | 68,903.54 | 2,167.98 | 62,156.25 | 202.1 |
| 8 | psvr | 9,148 | 55,658 | 263.81 | 0.55 | 4,240 | 22,243 | < 0.01% | < 0.01% | 4,245.78 | 15,940.34 | 3,336.74 | 8,662.02 | 170.7 |
| 9 | rift | 38,916 | 565,462 | 1,122.25 | 5.60 | 8,041 | 393,894 | < 0.01% | 0.15% | 18,099.32 | 275,490.06 | 14,773.80 | 246,702.20 | 170.1 |
| 10 | playstation | 74,484 | 1,578,155 | 2,147.95 | 15.64 | 9,912 | 707,453 | < 0.01% | 0.27% | 28,075.35 | 511,678.88 | 22,446.07 | 394,153.59 | 129.2 |
| 11 | augmented | 24,000 | 444,089 | 692.11 | 4.40 | 8,722 | 288,303 | < 0.01% | 0.11% | 13,494.35 | 198,818.19 | 12,849.52 | 166,792.39 | 128.3 |
| 12 | anxiety | 116,979 | 2,600,561 | 3,373.41 | 25.77 | 13,540 | 1,684,779 | < 0.01% | 0.64% | 51,483.89 | 1,212,964.75 | 44,086.77 | 1,067,579.88 | 126.0 |
| 13 | immersive | 34,553 | 745,551 | 996.43 | 7.39 | 9,883 | 545,505 | < 0.01% | 0.21% | 19,998.37 | 356,474.19 | 19,847.68 | 309,629.59 | 118.9 |
| 14 | alyx | 4,912 | 27,273 | 141.65 | 0.27 | 1,561 | 11,234 | < 0.01% | < 0.01% | 1,114.59 | 7,255.33 | 210.47 | 4,525.83 | 112.3 |
| 15 | cardboard | 15,732 | 404,785 | 453.68 | 4.01 | 4,459 | 287,238 | < 0.01% | 0.11% | 6,008.57 | 202,154.73 | 4,118.95 | 183,245.70 | 90.7 |

**Table 2.2**

*Top 15 Collocates—Modifier of "Virtual Reality" Nouns Modified by "Virtual Reality" "VR" & Prepositional Phrases (RQ2)*

| modifiers of "virtual reality" | | | nouns modified by "virtual reality" | | | prepositional phrases | | |
|---|---|---|---|---|---|---|---|---|
| Location-based | 108 | 5.2 | vr | 9 | 6.8 | ... of "virtual reality" | 6,865 | 8.7% |
| immersive | 628 | 5.1 | Bavor | 9 | 6.6 | ... in "virtual reality" | 5,110 | 6.4% |
| rift | 145 | 4.6 | Pov | 7 | 4.1 | ... for "virtual reality" | 2,168 | 2.7% |
| Vive | 90 | 4.3 | VR | 134 | 4.0 | ... to "virtual reality" | 1,277 | 1.6% |
| 360-degree | 147 | 4.1 | HTC | 9 | 2.8 | ... with "virtual reality" | 1,025 | 1.3% |
| Pictures | 121 | 4.0 | Ar | 14 | 2.0 | ... on "virtual reality" | 1,015 | 1.3% |
| Immersive | 51 | 3.9 | Sony | 9 | 0.6 | ... as "virtual reality" | 1,010 | 1.3% |
| location-based | 44 | 3.4 | | | | "virtual reality" in ... | 963 | 1.2% |
| VR | 276 | 3.4 | | | | ... into "virtual reality" | 895 | 1.1% |
| PC-based | 30 | 3.3 | | | | ... like "virtual reality" | 674 | 0.9% |
| room-scale | 27 | 3.2 | | | | "virtual reality" with ... | 477 | 0.6% |
| Oculus | 75 | 3.0 | | | | "virtual reality" for ... | 316 | 0.4% |
| cinematic | 128 | 2.5 | | | | ... about "virtual reality" | 287 | 0.4% |
| smartphone-based | 17 | 2.5 | | | | "virtual reality" to ... | 267 | 0.3% |
| Loungeroom | 15 | 2.4 | | | | "virtual reality" on ... | 228 | 0.3% |

**Table 3.3**

*Top 15 Collocates—Modifier of "VR," Nouns Modified by "VR," "VR" and/or, & Prepositional Phrases (RQ3)*

| modifiers of "VR" | | | nouns modified by "VR" | | | "VR" and/or ... | | | prepositional phrases | | |
|---|---|---|---|---|---|---|---|---|---|---|---|
| Gear | 33,040 | 10.9 | headset | 152,188 | 11.6 | Ar | 33,507 | 12.0 | ... in "VR" | 42,875 | 2.9% |
| PlayStation | 34,555 | 10.5 | goggle | 5,752 | 7.7 | rift | 3,614 | 9.2 | ... of "VR" | 40,804 | 2.8% |
| Oculus | 12,095 | 9.7 | Ar | 3,932 | 7.2 | Vive | 2,988 | 9.1 | ... for "VR" | 24,900 | 1.7% |
| PS | 5,776 | 8.6 | gaming | 7,411 | 7.1 | VR | 3,722 | 8.7 | ... to "VR" | 12,769 | 0.9% |
| Ar | 5,182 | 8.1 | hardware | 5,013 | 6.4 | headset | 3,355 | 8.6 | ... with "VR" | 10,200 | 0.7% |
| Playstation | 2,544 | 8.1 | experience | 53,596 | 6.3 | Quest | 1,309 | 7.8 | ... on "VR" | 8,695 | 0.6% |
| Skyrim | 2,156 | 7.9 | porn | 2,166 | 6.2 | cardboard | 1,339 | 7.7 | "VR" in ... | 7,978 | 0.5% |
| Arkham | 1,958 | 7.5 | arcade | 1,910 | 6.0 | Oculus | 888 | 7.4 | "VR" for ... | 5,445 | 0.4% |
| Superhot | 1,538 | 7.5 | glass | 4,463 | 6.0 | daydream | 751 | 7.3 | ... into "VR" | 4,997 | 0.3% |
| Samsung | 9,801 | 7.4 | demo | 2,044 | 5.9 | Mr | 899 | 7.2 | ... like "VR" | 3,773 | 0.3% |
| CEEK | 1,018 | 6.9 | controller | 3,907 | 5.9 | PSVR | 595 | 7.0 | "VR" on ... | 3,333 | 0.2% |
| Vive | 1,052 | 6.7 | content | 12,714 | 5.9 | gaming | 2,232 | 6.8 | ... about "VR" | 3,257 | 0.2% |
| daydream | 860 | 6.6 | simulation | 2,089 | 5.7 | IoT | 981 | 6.7 | ... as "VR" | 3,059 | 0.2% |
| sandbox | 994 | 6.5 | Worlds | 1,384 | 5.6 | reality | 3,099 | 6.6 | "VR" as ... | 2,591 | 0.2% |
| rift | 1,056 | 6.5 | bundle | 2,328 | 5.6 | metaverse | 565 | 6.6 | "VR" with ... | 2,163 | 0.2% |

**Figure 3.1**

*Top 15 Collocates of "Virtual Reality" by Grammatical Relation (RQ2)*

**Figure 3.2**
*Top 15 Collocates of "VR" by Grammatical Relation (RQ3)*

## Author Declarations

**Use of Generative AI**

In accordance with the World Association of Medical Editors (2023) guidelines for utilizing generative artificial intelligence (AI) in academic publications, the author discloses the following specific applications of generative AI in the research process:
1. Writing the abstract
2. Supporting the significance of research:
3. Integrating relevant literature: placing key words and frequency outputs within LLM to derive potential topical content

All prompt engineering inputs and generative AI outputs associated with this study are available for review on the project's Open Science Framework (OSF) page: https://osf.io/p3dwh/

**Credit Author Statement**

**Kwabena Yamoah**: Conceptualization, and Writing - Review and Editing.

**Cass Dykeman**: Methodology, Writing- Reviewing and Editing

**Funding Statement**

No external funding was received for conducting this research or for the preparation of this manuscript.

**Data Availability Statement**

The following are available on this research projects website: https://osf.io/p3dwh/ : (1) list of Sketch Engine sources, (2) data analyses, and (3) generative AI use logs.

**Conflicts of Interest Statement**

The authors declare that there are no conflicts of interest related to this work.

**Human Subjects Statement**

Given the public and published nature of the data, no human subjects reviewed was required.